\documentclass[10pt,twocolumn]{IEEEtran}
\emergencystretch=\maxdimen
\usepackage{booktabs}

\usepackage{subfigure}
\usepackage{amssymb}
\usepackage{amsfonts}
\usepackage{amsmath}
\usepackage{array}
\usepackage{cite}
\usepackage[dvips]{graphicx}
\usepackage{color}
\usepackage{comment}
\usepackage{makecell}
\usepackage{multirow}
\usepackage{textcomp,mathcomp}
\usepackage[linesnumbered, ruled, vlined]{algorithm2e}
\usepackage{caption}

\begin{document}
\newtheorem{definition}{\it Definition}
\newtheorem{theorem}{\bf Theorem}
\newtheorem{lemma}{\it Lemma}
\newtheorem{corollary}{\it Corollary}
\newtheorem{remark}{\it Remark}
\newtheorem{example}{\it Example}
\newtheorem{case}{\bf Case Study}
\newtheorem{assumption}{\it Assumption}
\newtheorem{property}{\it Property}
\newtheorem{proposition}{\it Proposition}

\newcommand{\hP}[1]{{\boldsymbol h}_{{#1}{\bullet}}}
\newcommand{\hS}[1]{{\boldsymbol h}_{{\bullet}{#1}}}

\newcommand{\ba}{\boldsymbol{a}}
\newcommand{\baq}{\overline{q}}
\newcommand{\bA}{\boldsymbol{A}}
\newcommand{\bb}{\boldsymbol{b}}
\newcommand{\bB}{\boldsymbol{B}}
\newcommand{\bc}{\boldsymbol{c}}
\newcommand{\bcD}{\boldsymbol{\cal D}}
\newcommand{\bcO}{\boldsymbol{\cal O}}
\newcommand{\bh}{\boldsymbol{h}}
\newcommand{\bH}{\boldsymbol{H}}
\newcommand{\bl}{\boldsymbol{l}}
\newcommand{\bm}{\boldsymbol{m}}
\newcommand{\bn}{\boldsymbol{n}}
\newcommand{\bo}{\boldsymbol{o}}
\newcommand{\bO}{\boldsymbol{O}}
\newcommand{\bp}{\boldsymbol{p}}
\newcommand{\bq}{\boldsymbol{q}}
\newcommand{\bR}{\boldsymbol{R}}
\newcommand{\bs}{\boldsymbol{s}}
\newcommand{\bS}{\boldsymbol{S}}
\newcommand{\bT}{\boldsymbol{T}}
\newcommand{\bw}{\boldsymbol{w}}

\newcommand{\balpha}{\boldsymbol{\alpha}}
\newcommand{\bbeta}{\boldsymbol{\beta}}
\newcommand{\bgamma}{\boldsymbol{\gamma}}
\newcommand{\bomega}{\boldsymbol{\omega}}
\newcommand{\bOmega}{\boldsymbol{\Omega}}
\newcommand{\bTheta}{\boldsymbol{\Theta}}
\newcommand{\bphi}{\boldsymbol{\phi}}
\newcommand{\btheta}{\boldsymbol{\theta}}
\newcommand{\bvarpi}{\boldsymbol{\varpi}}
\newcommand{\bpi}{\boldsymbol{\pi}}
\newcommand{\bpsi}{\boldsymbol{\psi}}
\newcommand{\bxi}{\boldsymbol{\xi}}
\newcommand{\bx}{\boldsymbol{x}}
\newcommand{\by}{\boldsymbol{y}}

\newcommand{\cA}{{\cal A}}
\newcommand{\bcA}{\boldsymbol{\cal A}}
\newcommand{\cB}{{\cal B}}
\newcommand{\cD}{{\cal D}}
\newcommand{\cE}{{\cal E}}
\newcommand{\cG}{{\cal G}}
\newcommand{\cH}{{\cal H}}
\newcommand{\bcH}{\boldsymbol {\cal H}}
\newcommand{\cK}{{\cal K}}
\newcommand{\cL}{{\cal L}}
\newcommand{\cM}{{\cal M}}
\newcommand{\cO}{{\cal O}}
\newcommand{\cR}{{\cal R}}
\newcommand{\cS}{{\cal S}}
\newcommand{\dcS}{\ddot{{\cal S}}}
\newcommand{\ds}{\ddot{{s}}}
\newcommand{\cT}{{\cal T}}
\newcommand{\cU}{{\cal U}}
\newcommand{\wt}[1]{\widetilde{#1}}

\newcommand{\mA}{\mathbb{A}}
\newcommand{\mE}{\mathbb{E}}
\newcommand{\mG}{\mathbb{G}}
\newcommand{\mR}{\mathbb{R}}
\newcommand{\mS}{\mathbb{S}}
\newcommand{\mU}{\mathbb{U}}
\newcommand{\mV}{\mathbb{V}}
\newcommand{\mW}{\mathbb{W}}

\newcommand{\uq}{\underline{q}}
\newcommand{\ubq}{\underline{\boldsymbol q}}

\newcommand{\red}[1]{\textcolor[rgb]{1,0,0}{#1}}
\newcommand{\gre}[1]{\textcolor[rgb]{0,1,0}{#1}}
\newcommand{\blu}[1]{\textcolor[rgb]{0,0,1}{#1}}

\title{SANNet: A Semantic-Aware Agentic AI Networking Framework for Multi-Agent Cross-Layer Coordination}

\author{\IEEEauthorblockA{Yong Xiao\IEEEauthorrefmark{1}\IEEEauthorrefmark{2}\IEEEauthorrefmark{3}, Haoran Zhou\IEEEauthorrefmark{1}, Xubo Li\IEEEauthorrefmark{1}, Yayu Gao\IEEEauthorrefmark{1},
Guangming~Shi\IEEEauthorrefmark{2}\IEEEauthorrefmark{4}\IEEEauthorrefmark{3}, Ping~Zhang\IEEEauthorrefmark{5}\\ 
\IEEEauthorblockA{\IEEEauthorrefmark{1} School of Elect. Inform. \& Commun., Huazhong Univ. of Science \& Technology, China}\\
\IEEEauthorblockA{\IEEEauthorrefmark{2} Peng Cheng Laboratory, Shenzhen, China}\\
\IEEEauthorblockA{\IEEEauthorrefmark{3} Pazhou Laboratory (Huangpu), Guangzhou, China}\\
\IEEEauthorblockA{\IEEEauthorrefmark{4} School of Artificial Intelligence, Xidian University, Xi'an, China}\\
\IEEEauthorblockA{\IEEEauthorrefmark{5} State Key Lab. of Networking \& Switching Tech., Beijing Univ. of Posts \& Telecom., Beijing, China}
}
\thanks{*This work is submitted to IEEE GLOBECOM'25. Copyright may be transferred without notice, after which this version may no longer be accessible.}
}

\maketitle

\begin{abstract}
Agentic AI networking (AgentNet) is a novel AI-native networking paradigm that relies on a large number of specialized AI agents to collaborate and coordinate for autonomous decision-making, dynamic environmental adaptation, and complex goal achievement. It has the potential to facilitate real-time network management alongside capabilities for self-configuration, self-optimization, and self-adaptation across diverse and complex networking environments, laying the foundation for fully autonomous networking systems in the future. Despite its promise, AgentNet is still in the early stage of development, and there still lacks an effective networking framework to support automatic goal discovery and multi-agent self-orchestration and task assignment. This paper proposes SANNet, a novel semantic-aware agentic AI networking architecture that can infer the semantic goal of the user and automatically assign agents associated with different layers of a mobile system to fulfill the inferred goal. Motivated by the fact that one of the major challenges in AgentNet is that different agents may have different and even conflicting objectives when collaborating for certain goals, we introduce a dynamic weighting-based conflict-resolving mechanism to address this issue. We prove that SANNet can provide theoretical guarantee in both conflict-resolving and model generalization performance for multi-agent collaboration in dynamic environment. We develop a hardware prototype of SANNet based on the open RAN and 5GS core platform. Our experimental results show that SANNet can significantly improve the performance of multi-agent networking systems, even when agents with conflicting objectives are selected to collaborate for the same goal. 
%
\end{abstract}


\section{Introduction}
\label{Section_Introduction}

With the fast proliferation of AI services and applications, next-generation communication networks will be dominated by a large number of diverse AI models and subsystems, coexisting, interacting, and communicating with one another\cite{Yang2022NetMagazine,XY2020Selflearning}. The existing data-transportation-focused networking architecture has faced unprecedented challenges to meet the fast-growing demand on communication and interaction needs of the future networks of AI agents.    
With the fast proliferation of AI services and applications, next-generation communication networks are expected to be characterized by a vast array of diverse AI models and subsystems that interact and communicate with one another. The existing data-transportation-centric networking architecture is increasingly confronted with novel challenges in accommodating the rapidly growing demands of real-time knowledge sharing, model coordination, collaborative decision-making, and adaptive resource orchestration, required by communication and networking of AI models and subsystems. 


Agentic AI networking (AgentNet) has attracted significant interest recently due to its potential to fundamentally address the limitations of the existing networking architecture by embracing an interactive learning and networking paradigm involving a large number of diverse AI agents with high-level autonomous, goal-driven, and adaptable decision-making capability\cite{Morris2024PositionAGI}. More specifically, instead of constructing a specific model for each individual task, agentic AI focuses on establishing and maintaining an interactive networking ecosystem where a diverse set of AI agents, each with unique knowledge, skillset and capabilities, and can autonomously communicate, interact and/or collaborate in archiving various complex goals across different environments\cite{Durante2024AgentAISurvey}.

Despite its promises, AgentNet is still in the early stage of development\cite{Shavit2023PracticesAgenticAI}. Most of the existing research focuses on multi-agent task planning and adaptation, ignoring the fact that the communication networking architecture plays a fundamental role in high-performance AgentNet systems. In particular, a recent study has already suggested that AgentNet is in essence a novel communication networking paradigm that supports effective communication and seamless interaction between human users and agents, as well as among multiple diverse agents within a system\cite{xiao2025AgentNet}. There is still a lack of a comprehensive framework that can automatically detect a user's semantic goal and self-orchestrate and coordinate different agents with different skillsets for fulfilling the detected goal\cite{ITU2023SAN}.

In this paper, we investigate the agentic AI system from the communication networking perspective. More specifically, we introduce a general multi-agent cross-layer coordination framework that supports the seamless interaction and collaboration of diverse agents associated with different layers of mobile networking systems, including the application-layer, network-layer, and physical-layer. We formulate the optimization problem for goal-oriented collaboration among agents associated with different layers and identify two major challenges faced by the multi-agent autonomous networking systems: (multi-agent) objective conflict and model generalization. The first one is that different agents, especially agents in layers, may have conflicting objectives when collaborating for a specific task, resulting in significant performance degradation and even divergence in goal achievement. The second challenge is that each agent can only observe a limited set of local data that cannot fully capture the complex real-world scenarios, resulting in a discrepancy between the predicted agent's performance and the final outcomes over the real-world data distributions. To solve these challenges, we first introduce two metrics, referred to as the C-error and G-error, to quantify the performance degradation caused by objective conflict and model generalization, respectively, and then propose a dynamic-weighting-based conflict-resolving mechanism that can provide a theoretical performance guarantee for both errors. We propose a novel semantic-aware agentic AI networking architecture, called SANNet, to allow the user's semantic goal to be learned, autonomously recognized, and separated into different subtasks, each of which can be fulfilled by an agent with the required skillset. We include our proposed conflict-resolving mechanism into the agent controller, so agents in different layers can collaborate in achieving the user's various semantic goals with guaranteed performance. Finally, we develop a hardware prototype based on the open RAN and 5GS core platform to evaluate the performance of SANNet in real-world scenarios. Experimental results show that SANNet can significantly reduce the C-error for multi-agent networks, compared to the existing state-of-the-art solutions.





\begin{figure}
\centering
\includegraphics[width=3.2 in]{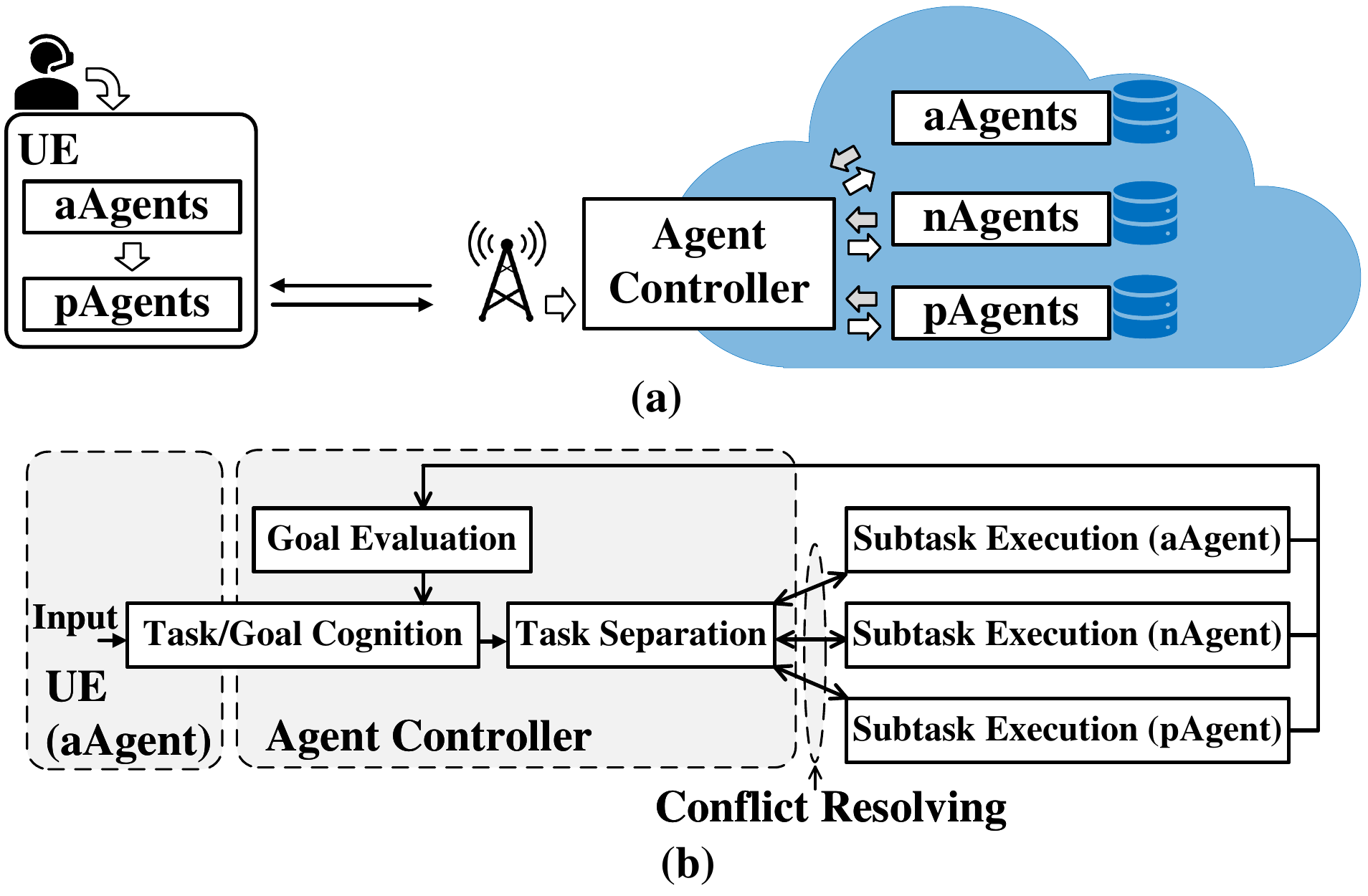}
\vspace{-0.1in}
\caption{(a) System model and (b) coordination procedures for a general multi-agent cross-layer mobile networking system.}
\label{Figure_SANNetModel}
\end{figure}


\section{System Model and Problem Formulation}
\label{Section_SystemModel}
\subsection{System Model}
We consider a general multi-task multi-agent cross-layer optimization problem for a mobile networking system, as illustrated in Fig. \ref{Figure_SANNetModel} (a). An agent is a physical or logical entity that can perceive the relevant users' semantic goals in specific environments and can learn and act autonomously to achieve the perceived goals. Each agent is composed of a single or a set of collaborative models designed with certain objectives and/or to meet a specific task demand. We consider a finite task space and let $\cM$ be the set of tasks that can be performed by a single agent or a set of agents.  
Agents deployed at different layers of the mobile system are generally designed to interact with different environments. 
In this paper, we focus on the cross-layer coordination and collaborative optimization involving multiple agents, and to simplify our description, we mainly focus on the following three types of agents. Our proposed solution, however, can be directly extended to more complex systems with more layers, as will be illustrated later.





\noindent{\em (1) Application-layer agent (aAgent)}: This corresponds to the agent that interacts with human users in various physical or virtual environments through some application interfaces. For example, virtual assistant applications installed on users' smart devices can interact with human users and infer the semantic goals of the users from the input of the UE. 
More formally, we define an aAgent by a tuple ${\cal G}^a = \langle \cA^a, \cS^a, \cL^a, \cD^a \rangle$ where $\cA^a$ is the set of acts that can be taken by aAgent, $\cS^a$ is the set of environmental state for an aAgent to sense and decide its act, $\cL^a$ is the set of task-related loss functions for aAgent to optimize, and $\cD^a$ is the set of data samples for training the aAgent's model. In this paper, we consider the AI model-based agentic AI networking system in which each agent focuses on training an AI model that can output a specific act based on the sensed state. Let $\omega^a$ be the model parameters associated with the aAgent. Generally speaking, aAgent cannot control the decision-making processes of other agents. Its performance, however, can be influenced by other agents' actions, especially those in the physical layer and network layer. We can write the loss function associated with task $m$ of aAgent as $l^a_m(\bOmega_m, \cD^a)$ where $\bOmega_m = \langle \bomega^a_m, \bomega^p_m, \bomega^n_m \rangle$ for $\omega^p$ and $\omega^n$ are model parameters associated with the agents in physical-layer and network-layer, respectively, as will be illustrated later. $\alpha^a_m \in \cA^a$, $s^a_m \in \cS^a$, and $l^a_m \in \cL^a$.

\noindent{\em (2) Physical-layer agent (pAgent)}: This corresponds to the agent that can interact and adapt to various physical-layer environments. For example, a pAgent deployed at the gNB or UE in a mobile network can estimate the spectrum availability and, in some cases, keep track of the channel state information (CSIs) connecting UE and gNB and make recommendations in channel selections or transmitter and receiver parameter choices. Similarly, we can also define a pAgent by a tuple ${\cal G}^p = \langle \cA^p, \cS^p, \cL^p, \cD^p \rangle$ where $\cA^p$ is the act space, 
$\cS^a$ is the environmental state space for the pAgent to estimate and decide its act, and $\cL^p$ is the set of task-related loss functions, and $\cD^p$ is the set of data samples for training the relevant models.
Similar to the aAgent, the decision-making process of the pAgent can also be performed by a model parameterized by $\omega^p_m$ that can output act decision based on the sensed physical-layer state information. We can write the loss function associated with task $m$ of pAgent as $l^p_m(\bOmega_m, \cD^p)$ for $\alpha^p_m \in \cA^p$, $s^p_m \in \cS^p$, and $l^p_m \in \cL^p$.

\noindent{\em (3) Network-layer agent (nAgent)}: This corresponds to the agent that interacts with the network-layer environments. For example, an nAgent deployed in the transportation layer or the core network of a mobile networking system can keep track of the routing and bandwidth resources between any network entities connecting the UE and its intended destination and adjust the routing and the amount of bandwidth resource accordingly. We define a nAgent by a tuple ${\cal G}^n = \langle \cA^n, \cS^n, \cL^n, \cD^n \rangle$ where $\cA^n$ is the set of acts that can be taken by nAgent, $\cS^n$ is the set of network-layer environmental states for an nAgent to estimate and decide its act, and $\cL^n$ is the set of task-related loss functions, and $\cD^n$ is the set of data samples for constructing the nAgent's model. Let $\omega^n_m$ be the model parameters associated with the nAgent. 
We can also write the loss function associated with task $t$ of pAgent as $l^n_m(\bOmega_m, \cD^n)$ for $\alpha^n_m \in \cA^n$, $s^n_m \in \cS^n$, and $l^n_m \in \cL^n$.

Each agent will not expose its local state or action to others. It can however report its capability via an agent-specific information tag, e.g., an agent card, to an agent controller and will only be activated when it has been called for a certain task.

\subsection{Problem Formulation}

Let us now describe the multi-agent cross-layer coordination procedure when a task request has been detected from the user's input as illustrated in Fig. \ref{Figure_SANNetModel} (b): (1) {\em (Semantic) goal detection:} A special aAgent can be deployed at the UE to keep track of the users' demand and recognize the user's semantics, e.g., a virtual assistant app can understand the user's semantics based on some key prompts using an LLM-based model; (2) {\em task separation:} The detected semantics will then be reported to an agent controller to decide the associated task as well as the corresponding subtasks that need to be performed in order to fulfill the user's semantic goal. Each subtask is assigned to a specific agent that has been previously exposed to its capability to the agent controller, e.g., if the user's demand involves initiating some data-heavy and high-bandwidth-demanding applications, pAgent and nAgent will be called to scan physical-layer resources such as spectrum availability and channel conditions, as well as network-layer routing and bandwidth resources; (3) {\em subtask execution:} Each agent, once received the assigned subtask, will make independent decisions, act accordingly, and report the result to the agent controller. (4) {\em goal (fulfillment) evaluation:} The user's goal will be fulfilled once all the agents assigned by the agent controller finish their duties. The aAgent on the UE side will continue to keep track of the user's input and repeat the above procedures when new demands are detected.

Since each agent is an independent decision maker, the agent controller cannot directly control its decision-making process. It can choose a specific set of agents when a certain task demand is detected. 
In this case, the local decision-making processes of the selected agents need to be carefully coordinated to make sure all the selected agents can achieve their objectives. We can define the multi-agent cross-layer optimization problem as follows:


\begin{eqnarray}
\mbox{\bf P1:}\; \min\limits_{\bOmega_m} L_{m}(\bOmega_m) := \langle l^a_m (\bOmega_m, \cD^a), l^p_m(\bOmega_m, \cD^p), l^n_m(\bOmega_m, \cD^n) \rangle. \nonumber
\end{eqnarray}
%

Note that the optimization objective of problem {\bf P1} involves a vector of objectives from different agents associated with different layers of the mobile system.  
These agents often have different or even conflicting objectives, and therefore it is generally impossible to find a single global optimal solution that can minimize the loss functions of all the agents.

Another major challenge faced by the multi-task multi-agent mobile systems is that the models developed by each agent need to be pre-trained to ensure responsive decision making, resulting in inaccurate and biased results when being called to act in new, unseen environment.





\section{SANNet Architecture}





We propose SANNet, a semantic-aware agentic AI networking framework that supports multi-agent cross-layer optimization on a new functional entity, the {\em agent controller}. 
The agent controller keeps track of the semantics of the user based on the input of the UE, e.g., language or visual expression of the user recorded by a UE aAgent. Each detected semantic demand can be associated with a set of different requirements associated with different layers of the system. A set of agents will then be selected to address various requirements raised from the users' demand. 
The agent controller will also evaluate the progress of task execution of various agents and mediate conflicts arising from divergent or conflicting objectives among them.

We provide a detailed description of the agent controller as follows. We will also present the theoretical analysis on the conflict resolution capability and generalization performance of the SANNet.



Let us first discuss the operational details of the agent controller as follows:


\noindent{\bf (1) Semantic cognition:} The semantics as well as the task/goal of the users can be recognized by the agent controller and linked to a specific task $t$ with a set of requirements for $t \in {\cT}$.

\noindent{\bf (2) Task separation and semantic translation:} The requirements associated with each task $t$ recognized by the agent controller will be translated into a set of subtasks, each is associated with an individual requirement raised by the users' demand. In this paper, we consider the cross-layer task separation, as illustrated in Fig. \ref{Figure_SANNetModel}. In this case, the requirements of each task $t$ recognized from the user's semantics will be translated into requirements associated with different layers, including application-layer, physical-layer, and network-layer.

\noindent{\bf (3) Agent selection:} Each agent will maintain a {\em agent card}, a meta-file consisting of the description of the agent's function, implementable environmental state, action space, associated layer, functional objective, and loss functions. Each agent will submit its agent card to the agent controller before it can be called. The agent controller will then map the requirements of each task $t$ in the application-layer, physical-layer, and network-layer to a set of agents associated with different layers. Each agent, once called, will sense its local environment and output the corresponding act under the sensed environment state. 




\noindent{\bf (4) Task evaluation and adjustment:} The agent controller will keep track of the subtask executions of different agents. Since different agents at different layers generally have different learning and optimization objectives, e.g., loss functions, when divergent objectives among some agents are detected, the agent controller will need to mediate and resolve the conflicts. In this paper, we propose a dynamic weighting-based conflict resolution solution that allows different agents to resolve conflicts and obtain a direction that optimizes all agents' objective functions jointly. 

The above operations will be repeated as long as new semantic demands of the users are detected. We assume that these operations cannot be interrupted once initiated, even if new demands are identified during their execution. 


Let us now describe how the agent controller can resolve conflicting objectives when agents associated with different layers have different goals, represented by different loss functions.
In the rest of this paper, we focus on the cross-layer collaboration among agents of different layers with conflicting objectives, i.e., the loss functions of different agents selected for performing specific tasks are different from each other.
%
It is known that in this case, there does not generally exist a single global optimal solution that minimizes all the agents' loss functions. 
Therefore, in this paper, we focus on the Pareto optimal solution where no agents can further improve their performance without making other agents worse off. More formally, we define the Pareto optimal solution for the multi-agent networking system as follows:

\begin{definition}
A solution profile $\bOmega$ is called {\em Pareto stationary} if there exists a set of non-negative weights $\gamma^a$, $\gamma^p$, and $\gamma^n$ summing to 1 such that $\sum_{i\in\{a, p, n\}}\gamma^i \nabla l^i_m (\bOmega^i_m, \cD^i)$. A solution profile $\bOmega^*$ is {\em Pareto optimal} if no other Pareto stationary solution $\bOmega$ for $\bOmega \neq \bOmega^*$ such that $l^i_m (\bOmega) \le l^i_m (\bOmega^*, \cD^i)$ for all $i \in \{a, p, n\}$ and $l^i_m (\bOmega) < l^i_m (\bOmega^*, \cD^i)$ for at least one $i \in \{a, p, n\}$.  
\end{definition}

We further introduce the following performance metric to quantify the error caused by the conflicting objectives among agents, and we will then propose solutions to minimize this error. 

\noindent{\bf (Muti-agent) conflicting error (C-error):} This corresponds to the error caused by the conflicting gradients among different agents, defined as follows: 
\begin{eqnarray}
    \cE_C := \| \sum_{i\in\{a, p, n\}}(\gamma^i-\gamma^{i*}) \nabla l^i_m (\bOmega_m, \cD^i) \|,
\end{eqnarray}

Previous results \cite{desideri2012multiple} have already proved that the optimal solution achieves the Pareto stationary solution minimizes the convex combination of all agents' losses, i.e., we can therefore rewrite the optimization problem {\bf P1} into the following problem: 
%
\begin{eqnarray}
\mbox{\bf P2:} \;\;\;  \min\limits_{\bOmega_m} \| \sum_{i\in \{a, p , n\}} \gamma^i \nabla l^i_m (\bOmega_m, \cD^i) \|,
\end{eqnarray}
where $\gamma^i$ satisfies $\sum_{i\in \{a, p , n\}} \gamma^i = 1$. 

In this paper, we propose a dynamic weighting-based conflict resolving solution to minimize the C-error. 
In the rest of this section, we first introduce the dynamic weighting algorithm for the agent controller to dynamically control the weights among agents to influence the convergence directions. We will prove that our proposed algorithm can minimize the C-error and accelerate the convergence of the multi-agent decision-making process for cross-layer optimization.

Let us first introduce the dynamic weighting algorithm as follows: Generally speaking, the agent controller should always estimate the gradient descent directions that improve the losses of all the agents. In practice, however, calculating the full-batch gradients can be costly and therefore we adopt a stochastic learning solution in which the gradients of agents have been replaced by the stochastic approximated versions sampled at a single data sample $d^i_{m}$ for each agent $i$ for $i\in \{a, p, n\}$. 
The gradient directions as well as the model parameters of all three layers of agents are updated iteratively with independently sampled data.
More specifically, at each iteration $t$, we draw three independent data samples $\{d_{m,t(j)}^i\}_{j\in \{1,2,3\}}$ from $\cD^i$ in parallel for different agents and perform the following updates on weighting factors of gradient directions and agents' model parameters, respectively, as follows:
\begin{eqnarray}
    \gamma_{t+1}^i &=&  \gamma_t^i - \eta_t \nabla l^i(\bOmega_t, \gamma_t^i, {d^i_{t,1}})^{\top} \nabla l^i(\bOmega_t, \gamma_t^i, {d^i_{t,2}}), \label{gamma}\\
    \bOmega_{t+1} &=& \bOmega_t - \beta_t \nabla l^i(\bOmega_t, \gamma_{t+1}^i, {d^i_{t,3}}), \label{omega}
\end{eqnarray}
where $\eta_t, \beta_t$ are step sizes of the gradient direction and agents' model parameter update. The detailed procedures are illustrated in Algorithm 1. 

\begin{algorithm}[t]\label{Algorithm_DynamicWeight}
\caption{Dynamic Weighting-based Conflict Resolving Mechanism}
\KwIn{Training datasets $\cD^a, \cD^p, \cD^n$; initial model $\bOmega_0$; initial weight $\gamma^a, \gamma^p, \gamma^n$.}
\KwOut{Task-specific agents $\bOmega_m = [\bomega^a_m, \bomega^p_m, \bomega^n_m]$.}
\While{not converged}
{
    \For{$ i \in \{a, p, n\}$ \emph{\textbf{parallel}}}
    {
    Compute dynamic weight $\gamma_{t+1}$ by (\ref{gamma});\\
    Update agents' parameters by (\ref{omega});
    }
}
  \Return Trained agents $\bOmega_m$ for specific task
\end{algorithm}

We can prove the following bound on the C-error of Algorithm 1.

%
\begin{theorem}
\label{Theorem_UpperBoundCError}
Suppose the following assumptions holds: \emph{(i)} $\nabla l^i (\bOmega, \gamma)$ is $\ell'_f$-Lipschitz continuous for any data sample; \emph{(ii)} $l^i (\bOmega, \gamma)$ is $\ell_f$-Lipschitz continuous for any data sample.
    Then, the following upper bound holds for Algorithm 1:   
    \begin{eqnarray}
        \cE_C \le \frac{4}{\eta T} \!+\! 6 \sqrt{3 \ell'_{f}\ell_{f}^2\frac{\beta}{\eta}} \!+\! 3\eta \ell_f^4.
        \label{eq_Theorem1_Ec}
    \end{eqnarray}
\end{theorem}
\begin{IEEEproof}
Due to the limit of space, we provide a sketch of proof of the above theorem as follows. According to Lemma 18 in \cite{chen2023three}, there always exists a positive constant $\rho$ such that the following holds
\begin{eqnarray}
    \cE_C \le \rho + \frac{4}{\eta T}(1+\rho^{-1}\beta T C_1) + \eta C_2,
\end{eqnarray}
where $C_{1} =\frac{1}{T} \sum_{t=0}^{T-1} \mathbb{E}\|\nabla L(\bOmega_{t+1}) + \nabla L(\bOmega_{t})\| \cdot \| \sum_{i\in \{a, p , n\}} \nabla l^i(\bOmega_t, \gamma_{t+1}^i, {d^i_{t,3}})\|$ and $C_{2} = \frac{1}{T} \sum_{t=0}^{T-1} \mathbb{E} \| \sum_{i\in \{a, p , n\}} \nabla l^i(\bOmega_t, \gamma_t^i, {d^i_{t,1}})^{\top} \nabla l^i(\bOmega_t, \gamma_t^i, {d^i_{t,2}}) \|^{2}$.

According to the Lipschitz property, $C_1$ and $C_2$ are bounded by $6\ell'_{f}\ell_{f}$ and $3\ell_{f}^2$, respectively. By setting $\rho = 2 \sqrt{3 \beta \ell'_{f}\ell_{f}^2/\eta}$, we can obtain the result in (\ref{eq_Theorem1_Ec}). 
\end{IEEEproof}

We can observe that, by setting $\beta = \Theta (T^{-3/4})$ and $\eta = \Theta (T^{-1/4})$ in Theorem \ref{Theorem_UpperBoundCError}, the C-error converges to the conflict-resolving direction at rate $\mathcal{O}(T^{-1/4})$.



In addition to the C-error, let us also evaluate the agent's generalization capability when being deployed in the open dynamic environment that cannot be fully captured by its local model training dataset. Specifically, we adopt a commonly adopted metric to quantify the generalization capability of each agent, defined as follows:

\noindent{\bf (Model) generalization error (G-error):} We define the generalization error as the discrepancy between the agent deployment performance based on the real data distribution and the training performance obtained on the training dataset. More formally, let ${\tilde l}^{i}_m (\omega_m^i)$ be the population loss of the agent $i$'s model for $i\in\{a, n, p\}$, evaluated over the real data distribution of task $m$. We can then define G-error of agent $i$'s model $\cE_G^i$ as the difference of the gradients between the population loss ${\tilde l}^{i}_m (\omega_m^i)$ calculated based on the real distribution of agent deployment and the empirical loss $l^i_m(\omega_m^i, \cD^i)$ obtained based on the training dataset $\cD^i$, i.e., the G-error of agent $i$ $\cE_G^i$ can be written as:
\begin{eqnarray}
    \cE_G^i = \| \nabla l^i_m(\omega_m^i, \cD^i) - \nabla {\tilde l}^{i}_m (\omega_m^i) \|,
\end{eqnarray}

We can also define the overall generalization error of all the agents selected to solve task $m$ as follows:
\begin{eqnarray}\label{E_G}
    \cE_G = \| \sum_{i\in\{a, p, n\}}\gamma^i (\nabla l^i_t (\bOmega_t, \cD^i) - \nabla {\tilde l}^{i}_t (\bOmega_t) ) \|
\end{eqnarray}

We can then prove that the conflict-resolving mechanism can have the following bound on the G-error. 
\begin{theorem}
\label{Theorem_UpperBoundGError}
Suppose the Frobenius norm of the summation of gradients of all agents is upper bounded by constant, i.e, $\mathbb{E}[\| \sum_{i\in \{a, p, n\}} \gamma^i \nabla l^i_m (\bOmega_m, \cD^i) \|_\text{F}^2] \le U^2$ for $U$ is a constant. Then, the G-error of Algorithm 1 is upper bounded by $\cE_G \le \cO(T^{\frac{1}{2}}D^{-\frac{1}{2}})$.
\end{theorem}

\begin{IEEEproof}
We provide a sketch of proof of the above theorem as follows. We can first prove that, if a single training data sample is modified or removed, e.g., the training dataset $\cD^i$ of agent $i$ is changed to ${\tilde \cD}^i$ where ${\tilde \cD}^i$ is identical to ${\tilde \cD}^i$ except that a single data sample is different, the gradient of the loss function in our proposed Algorithm 1 is upper bounded by a constant $\sup \ \mathbb{E} [ \|  \sum_{i\in\{a, p, n\}} \gamma^i (\nabla l^i_t (\bOmega_t, \cD^i) - \nabla l^i_t (\bOmega_t, {\tilde \cD}^i))\|^2_\text{F}] \le \epsilon^2$. 

We can then follow the same line as \cite{chen2023three} and prove the G-error has the following upper bound $\cE_G \le 4 \epsilon + \sqrt{\frac{V}{D}}$, 
where $V$ is the variance of gradients and $D$ is the size of the training dataset.
Substituting the upper bounds of gradients assumed in Theorem \ref{Theorem_UpperBoundGError}, we can also prove that $\epsilon$ is bounded by a constant, given by $\epsilon^2 \le \frac{4U^2T}{D}$. Finally, by substituting this bound of $\epsilon$ into the upper bound of $\cE_G$, we can obtain the result.
\end{IEEEproof}

\vspace{-0.1in}
\section{Prototype and Experimental Results}


\subsection{Prototype}
\begin{figure}
\centering
\includegraphics[width=3.3 in]{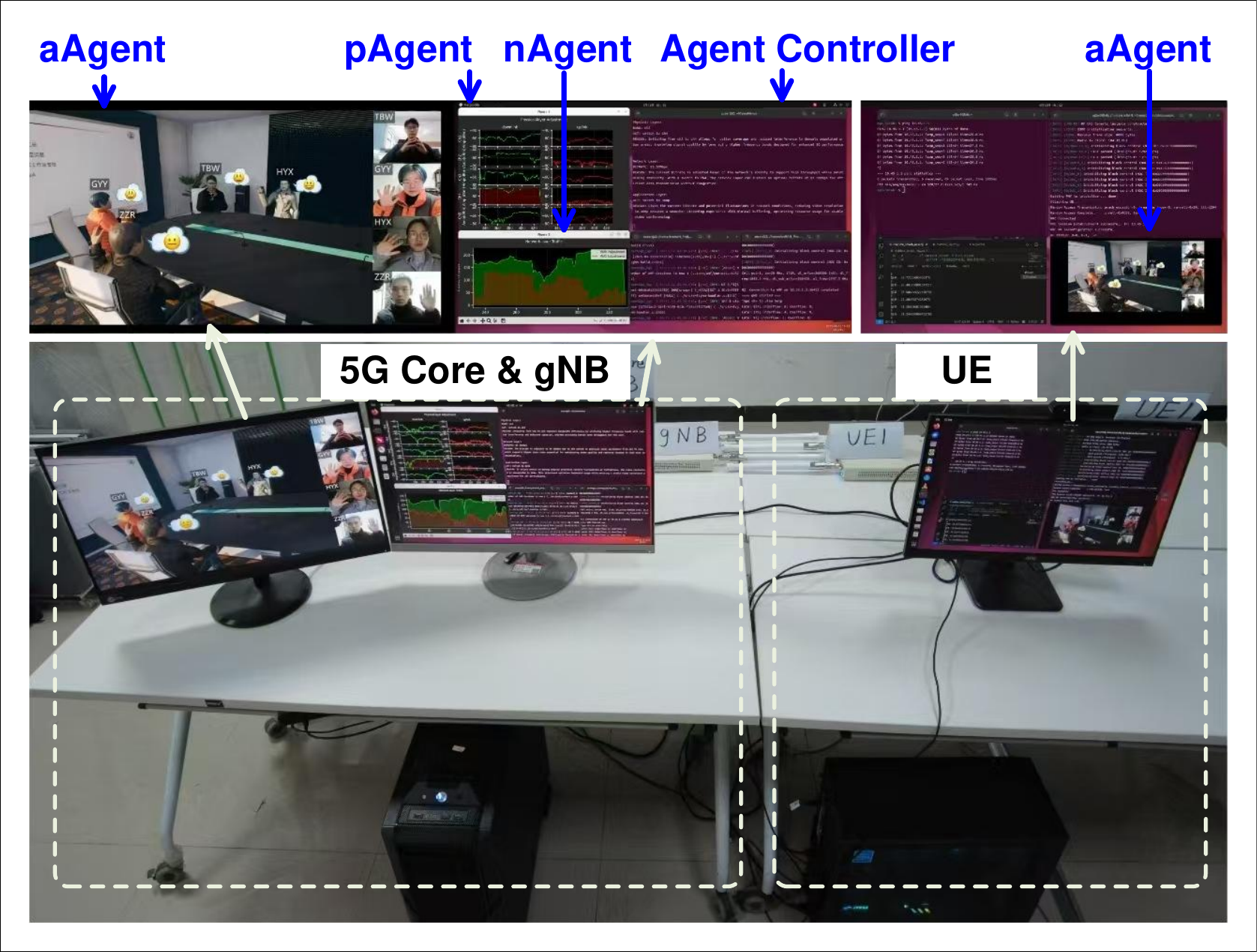}
\caption{A SANNet prototype.}
\vspace{-0.2in}
\label{Figure_Prototype}
\end{figure}

We develop an open RAN and open 5GS-based SANNet prototype for evaluating the performance of our proposed solution in various practical environments, as shown in Fig. \ref{Figure_Prototype}. The detailed hardware and software platforms implemented in our prototype are described as follows:

\noindent{\bf Hardware platform:} The hardware of our SANNet prototype is composed of a UE, a gNB, and a 5G core network. The UE is an NI 2944R USRP connected to a desktop computer installed with an Intel(R) Core(TM) i7 CPU@2.4 GHz with 32 GB memory and 1TB SSD. The gNB is an NI 2944R USRP connected to a workstation computer installed with an Intel(R) Core(TM) i9-13900K CPU@5.8GHz, 128.0GB RAM@4000.0MHz, 1 TB SSD, 4 TB HDD, and 1 NVIDIA GeForce RTX 4090 GPU. 

\noindent{\bf Software platform:} The software of our prototype consists of an open source 5G open RAN compatible software radio platform, srsRAN, installed on both computers of the UE and the gNB\cite{Habibi2024AIMLinORAN}. An open 5GS-based 5G core network is also installed on the workstation of gNB.

\subsection{Experimental Setup}
We consider immersive communication driven by user QoE as a case study to demonstrate how SANNet can infer the implicit semantics of the user and autonomously initiate cross-layer joint optimization between aAgent, nAgent, and pAgent to meet the inferred QoE demand of the user.
In our case study, an aAgent, e.g., a virtual assistant application, detects a user's unsatisfactory with the quality (resolutions) of an online streamed video by detecting one or multiple prompts from the user's language. For example, if the aAgent detects prompts such as ``increase video resolution" and ``make video clearer" from the user's language, it will understand that the semantic goal of the user corresponds to ``increasing the resolution of an online streamed video without sacrificing other QoE-related performance such as smoothness and latency of the video streaming services". In this case, the aAgent will send this detected goal to the agent controller which corresponds to an LLM that can understand user's language, infer semantic goal, and separate the task goal into three subtasks to be assigned to three different agents: another aAgent which corresponds to a Unity-based 3D video rendering application that can adjust video resolutions among 360p, 480p, 640p, 720p, and 1080p; a pAgent which corresponds to a multi-channel sensing and assigning agent that can sense the CSIs of multiple channels and estimate the data rates that can be supported by each channel; and nAgent which corresponds to a network bandwidth tracking and prediction agent that can keep track of the network bandwidth, max data traffics, between the IP addresses of the UE and 5G core.

To evaluate the multi-agent multi-objective sensors of the above scenario, we built a multi-channel dataset and trained three transformer-based prediction models for aAgent, pAgent, and nAgent, respectively. We describe our dataset, configurations of different agents' models, and that of the agent controller as follows:

\noindent{\bf Dataset:} We build a multi-channel dataset consisting of data samples collected by aAgent, pAgent, and nAgent in application-layer, physical-layer and network-layer. More specifically, for the dataset of aAgent, we assume the user can only request changes of video quality every 5 seconds and randomly generate a time sequence of the user's requests. For the dataset of pAgent, we record CSI data samples in five 5G NR frequency bands, including n1, n2, n3, n5, n7 bands. For the dataset of nAgent, we record the maximum achievable bandwidth between IP addresses of the UE and 5GS core measured by the iPerf3 tool.

\noindent{\bf Agents:} We train three time series prediction models based on the transformer network with three different loss functions $l_1$ loss, mean square error (MSE) loss, and log-Cosh loss for the sensing module of aAgent, pAgent, and nAgent, respectively. Each agent will feedback its prediction result to the agent controller and coordinate with each other following the aforementioned steps.

\noindent{\bf Agent controller:} We adopt an open-source AI agent platform, OpenManus, as the agent controller and install an open-source transformer-based LLM model, Qwen-7B, to infer the user's QoE via language input.


\subsection{Experimental Result}




\begin{figure}
\begin{minipage}[t]{0.5\linewidth}
\includegraphics[width=1.8 in]{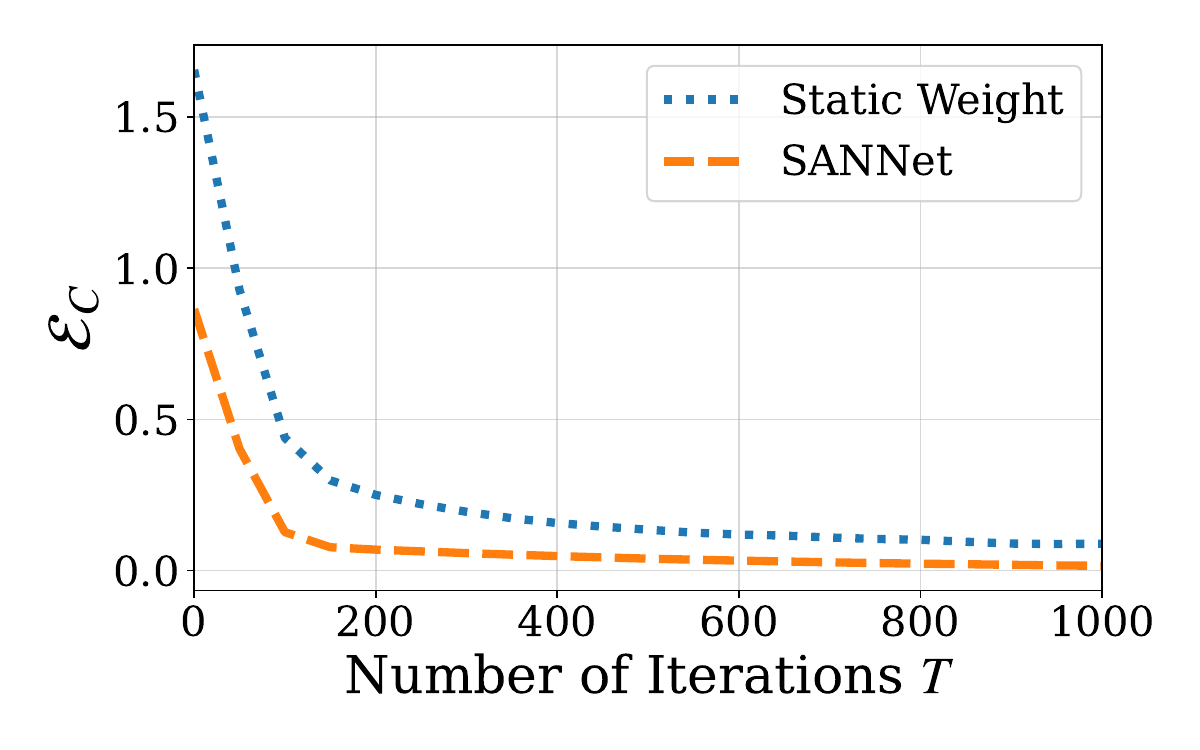}
\vspace{-0.2in}
\caption*{(a)}
\label{Fig_ExperimentCerror}
\end{minipage}
\begin{minipage}[t]{0.45\linewidth}
\includegraphics[width=1.8 in]{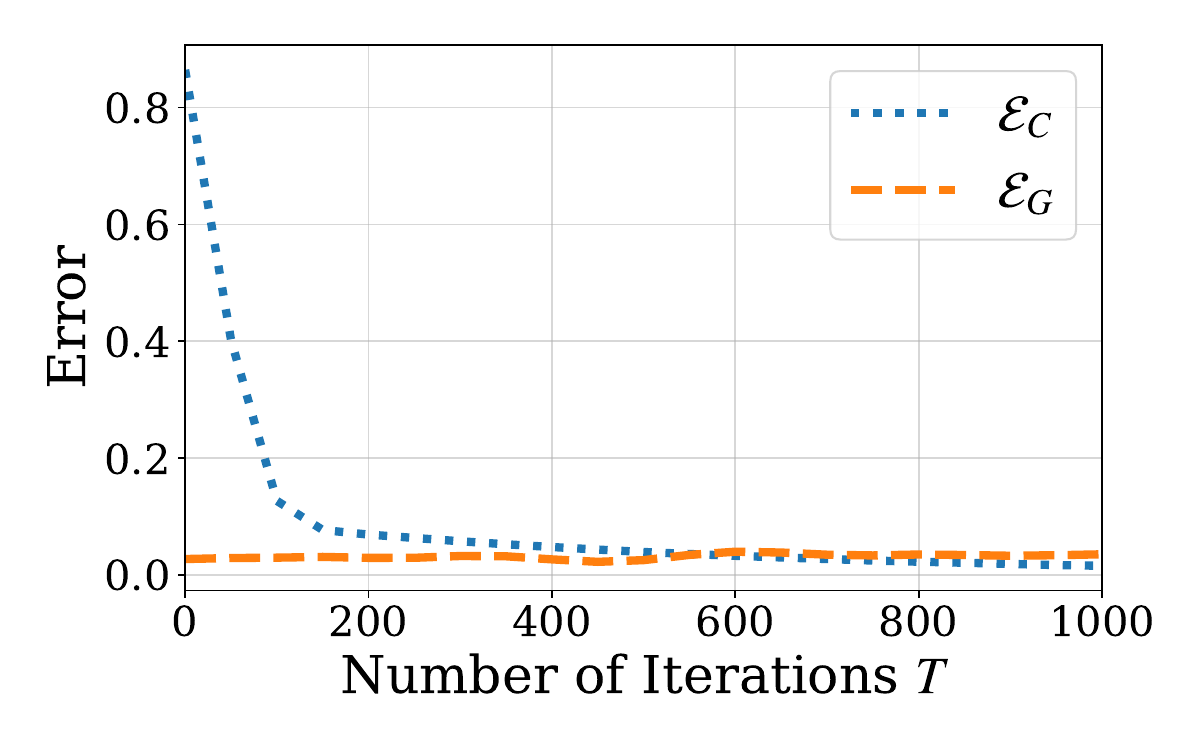}
\vspace{-0.2in}
\caption*{(b)}
\label{Fig_ExperimenrCGerrorTradeoff}
\end{minipage}
\vspace{-0.1in}
\caption{\small (a) C-error of our proposed dynamic weight-based conflict resolving mechanism, compared to the static weight solution, and (b) a tradeoff between C-error and G-error.}
\vspace{-0.1in}
\label{Figure_ExperimenrCandGerror}
\end{figure}

We first evaluate the conflict resolving performance of our proposed dynamic-weighting mechanism for the agent controller. As mentioned earlier, three different agents adopt three different loss functions, representing three different objectives, resulting in degraded performance for model training. In Fig. \ref{Figure_ExperimenrCandGerror}(a), we present the C-error of the models trained under different numbers of iterations, compared to the traditional multi-agent collaboration model training solution with static weight. We can observe that our proposed dynamic weight-based solution can reduce up to 63\% of C-error, compared to the static weight approach.

We also compare the G-error and C-error under different numbers of iterations in Fig. \ref{Figure_ExperimenrCandGerror}(b). We can observe that, compared to the C-error, which decreases dramatically, the G-error increases slightly when the number of iterations becomes large. This is because when the number of iterations becomes large, the model tends to fit too closely to the training dataset, resulting in decreased performance when the ground truth environment is different. 








\vspace{-0.1in}
\section{Conclusion}
This paper proposes SANNet, a novel semantic-aware agentic AI networking architecture that autonomously identifies the user's semantic goal and autonomously divides the identified goal into different subtasks for different agents. We introduce a novel functional entity, the agent controller, employed with a novel dynamic weighting-based conflict-resolving mechanism. We develop a hardware prototype and our experiment result suggests that SANNet significantly improves the performance of multi-agent networking systems. 
\vspace{-0.1in}

\bibliographystyle{IEEEtran}
\bibliography{DeepLearningRef}

\end{document}